\pdfoutput=1

\documentclass[11pt]{article}

\usepackage[final]{acl}

\usepackage{times}
\usepackage{latexsym}
\usepackage[T1]{fontenc}
\usepackage[utf8]{inputenc}
\usepackage{microtype}
\usepackage{inconsolata}
\usepackage{hyperref}
\usepackage{url}
\usepackage{booktabs}
\usepackage{amsfonts}
\usepackage{nicefrac}
\usepackage{fancyhdr}
\usepackage{graphicx}
\usepackage{subfigure}
\usepackage{multicol,multirow}
\usepackage[ruled,linesnumbered]{algorithm2e}
\usepackage{xcolor}
\usepackage{ifpdf}
\usepackage{arydshln}
\usepackage{moreverb}
\usepackage{amsmath}

\title{Domain Lexical Knowledge-based Word Embedding Learning for Text Classification under Small Data}

\author{
Zixiao Zhu\textsuperscript{1} \quad \textbf{Kezhi Mao\textsuperscript{1}\thanks{Corresponding author}} \\
\textsuperscript{1}School of Electrical and Electronic Engineering, Nanyang Technological University, Singapore \\
\texttt{\{zixiao.zhu, ekzmao\}@ntu.edu.sg}
}

\begin{document}
                
\maketitle

\begin{abstract}
Pre-trained language models such as BERT have been proved to be powerful in many natural language processing tasks. But in some text classification applications such as emotion recognition and sentiment analysis, BERT may not lead to satisfactory performance. This often happens in applications where keywords play critical roles in the prediction of class labels. Our investigation found that the root cause of the problem is that the context-based BERT embedding of the keywords may not be discriminative enough to produce discriminative text representation for classification. Motivated by this finding, we develop a method to enhance word embeddings using domain-specific lexical knowledge. The knowledge-based embedding enhancement model projects the BERT embedding into a new space where within-class similarity and between-class difference are maximized. To implement the knowledge-based word embedding enhancement model, we also develop a knowledge acquisition algorithm for automatically collecting lexical knowledge from online open sources. Experiment results on three classification tasks, including sentiment analysis, emotion recognition and question answering, have shown the effectiveness of our proposed word embedding enhancing model. The codes and datasets are in \url{https://github.com/MidiyaZhu/KVWEFFER}.
\end{abstract}

\section{Introduction}
\label{sec1}
Bidirectional Encoder Representation from Transformers (BERT) has proved to be powerful in many NLP tasks due to its capability of capturing contextual information \citep{devlin2018bert}. 
However, BERT also has some limitations. It is found that BERT lacks domain-specific knowledge \citep{yan2021sakg,liang2023multi,mutinda2023sentiment}, which may hinder its performance in applications where domain-specific knowledge plays critical roles. It was also found that BERT could map sentiment words with opposite polarity to similar embedding \citep{zhu2023knowledge}. Since sentiment words are keywords in sentiment analysis, the above issue hiders discriminative feature learning for classification tasks \citep{rezaeinia2019sentiment}. In addition to the issue of similar embedding of opposite polarity keywords, we found that BERT embedding may also have the following two issues: i) the embedding of sentiment words of the same polarity could be very different, lacking within-class cohesion; ii) the embedding of sentiment words lacks class-discriminative power in mutual assistance. Both issues are detrimental to pattern classification. We conducted extensive experimental studies and found that the above issues often happen in the following scenarios. First, the contexts of keywords with opposite polarity are similar. Second, the contexts of the keywords are noisy, containing limited information relevant to the class labels. Third, the contexts are very short. Our analysis discovers that the above issues are due to the context-based learning of BERT embedding, though context-based learning is the main merit of BERT leading to its success in many NLP tasks. To address this side-effect of context-based learning, we believe the word embedding should comprise two parts: one part captures contextual information just as BERT embedding, and the other part should contain class discriminant information less dependent on contexts. To produce class-discriminative word embedding less dependent on contexts, in this paper, we investigate the use of domain-specific lexical knowledge, instead of training data, to build an embedding enhancement model to map BERT embedding to a new discriminative space where the within-class cohesion and between-class separation are maximized.

Knowledge infusion is vital for enhancing domain-relevant learning \citep{khan2023sentiment}.
While retraining or fine-tuning BERT with domain-specific knowledge is resource-intensive \citep{sun2020ernie} and often ineffective with small datasets \citep{mutinda2023sentiment}(i.e., < 10K examples \citep{zhangrevisiting}), retrofitting word embeddings with auxiliary knowledge presents a simpler, more efficient alternative. However, this approach comes with its own challenges, such as gaps in coverage when certain words are not included in the knowledge lexicon \citep{cui2021knowledge}, and the potential loss of contextual nuances captured by pre-trained models \citep{biesialska2020enhancing}. To address these issues, we develop a knowledge acquisition algorithm to gather domain-specific lexicons from online open sources. Using this lexical knowledge, we propose a novel embedding learning model to map any word embeddings, including unseen ones, into a discriminative space. These enhanced embeddings are combined with BERT embeddings, boosting the effectiveness of feature learning. Unlike traditional retrofitting, our method works independently as an auxiliary component, offering greater flexibility and enriching the classifier with domain-specific information to improve discriminative performance.

The main contributions of this paper can be summarized as follows:
\begin{enumerate}
\item We investigate the side-effect of context-based learning on BERT embedding, and develop a lexical knowledge-based word embedding learning model to map BERT word embedding in a new discriminative space to achieve within-class cohesion and between-class separation.
\item We develop a lexical knowledge acquisition algorithm that can automatically acquire class-specific lexicon from various open resources.
\item We conduct extensive experiments and analysis on three text classification tasks, which show that the proposed method produces state-of-the-art results in sentiment analysis, emotion recognition and question answering.
\end{enumerate}

\section{Related Work}
\label{sec2}
Pre-trained language models (PLMs) such as BERT are widely used in various natural language processing tasks. However, such a generic language model may not fit all tasks well \citep{sun2019utilizing}. Retraining a domain-oriented language representation model needs a vast textual training corpus to achieve optimal results \citep{sun2020ernie,ji2021survey}. Fine-tuning strategies, though often employed to adapt these models to specific tasks, require substantial training data and might lead to instability \citep{mosbach2020stability} or overfitting \citep{kamyab2021attention}. 

One promising solution lies in knowledge-enhanced methods that infuse task-focused knowledge into the classification model \citep{mar2020s}, potentially outperforming traditional fine-tuning techniques by generating more discriminative features \citep{wang2023hierarchically,zhao2022kesa}. The knowledge can be categorized as data-oriented or lexicon-oriented. The dataset-oriented knowledge like class label \citep{zhang2021improving}, topic \citep{li2020context}, or position \citep{ishiwatari2020relation} can be incorporated within the feature learning to enhance task-oriented feature attention. However, it solely relies on the data, without incorporation of external knowledge, leading to limited performance improvement, especially when the training data is small. 

The lexicon-oriented knowledge-enhanced methods focus on enriching pre-trained word embeddings through alignment with external domain-specific lexicons \citep{khan2023sentiment}. This process of refinement is generally more cost-effective than training a new language model from scratch \citep{zheng2022using}. To address the issue of "blind spots" in refined embeddings for unseen words beyond the employed lexicon, several studies \citep{ wang2022contextual, cui2021knowledge, vulic2018post} use mapping functions to apply lexicon semantic features to all words. Nevertheless, these transformed embeddings, being generated from text and containing additional contextual information, may not correspond accurately to the individual words in the lexicon \citep{colon2021retrogan}. Furthermore, this transformation might occasionally undermine the pre-trained contextual information \citep{glavavs2018explicit}. To address the limitations, we introduce a novel approach that synergizes external knowledge and pre-trained word embedding. This strategy employs domain-specific lexical knowledge to transform BERT word embeddings into a more discriminative space. This lexical knowledge-based embedding learning builds on single-word BERT embedding and word-level lexicon. More importantly, it does not demand BERT model fine-tuning or new language model development, marking it a computationally efficient solution. 

\begin{figure*}[!t]
\centering  
\subfigure[The word representation vectors of sentiment words in opposite polarities have high embedding similarity due to the similar contexts (sourced from \citep{mohammad2017wassa}).]{
\label{simliar context}
\includegraphics[width=0.95\textwidth]{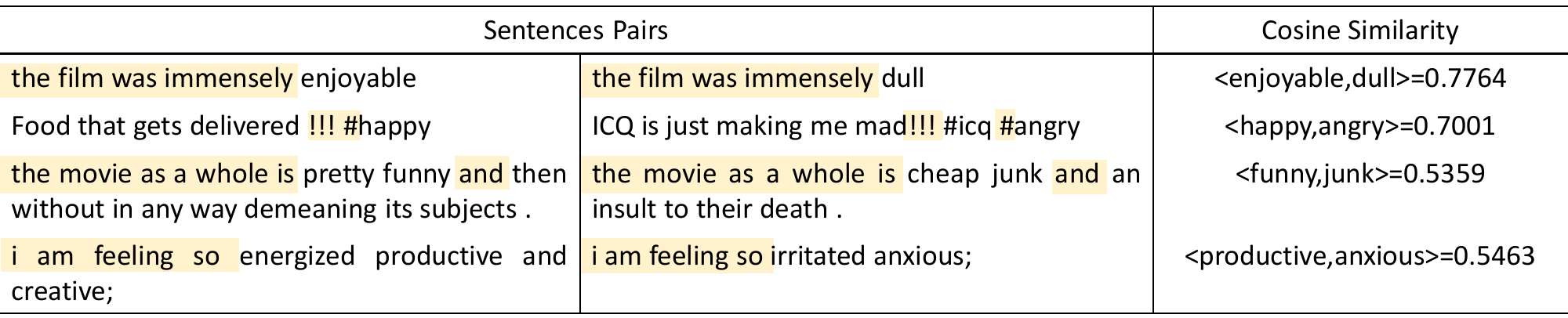}}
\subfigure[The word representation vectors of sentiment words in the same polarity show low similarity even under keywords-relevant contexts (sourced from \citep{mohammad2017wassa}). ]{
\label{Different context}
\includegraphics[width=0.95\textwidth]{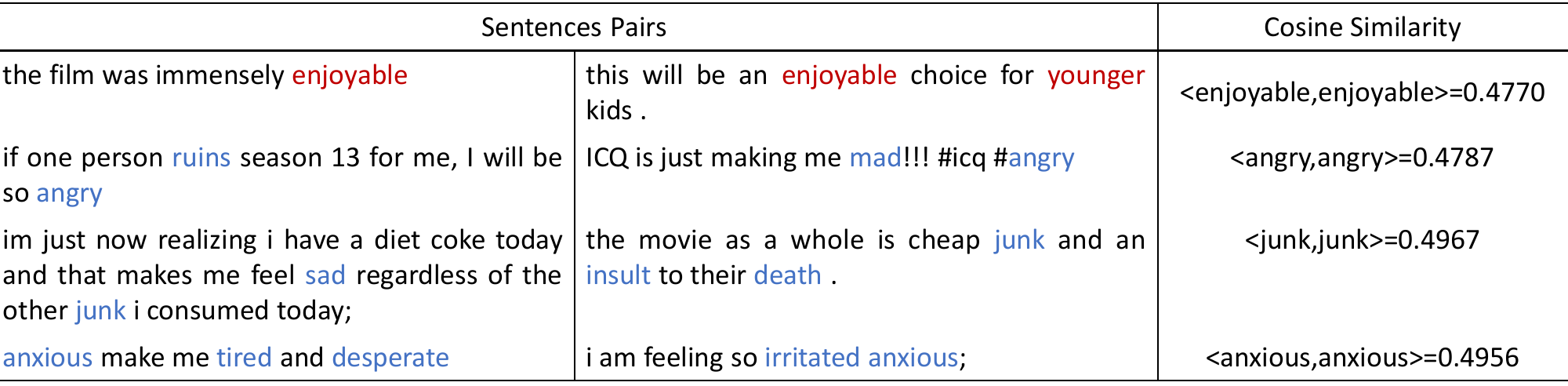}}
\subfigure[The word representation vectors of sentiment words (sourced from \citep{liu2005opinion}) generated by singly inputted into BERT have a higher or close average between-class cosine similarity (0.6685) than that of the within-class sentiment words (0.6668 for positive words and 0.6881 for negative words). ]
{\label{No context}
\includegraphics[width=0.95\textwidth]{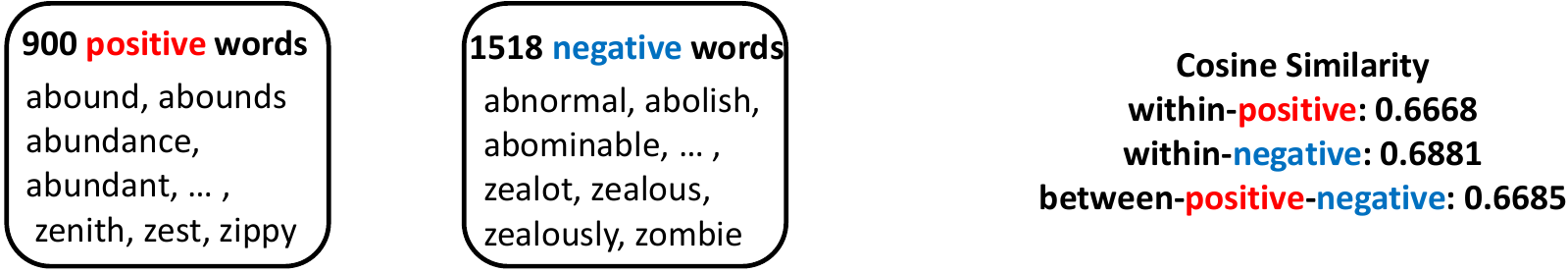}}
\caption{Exploring the Impact of Context-Based Learning in BERT on Sentiment Words: We present the cosine similarities between sentiment word embeddings from BERT across (a) similar contexts, (b) contexts containing sentiment-related keywords, and (c) inherent features.}
\end{figure*}

\section{Method Description}
\subsection{Problem statement}
As aforementioned, BERT embedding could have the following issues:

\noindent\textbf{Wrong Context-Based Discriminative Feature of Sentiments Polarities:} BERT embeddings can paradoxically show greater similarity between words of opposite sentiment polarity than between words of the same sentiment, especially when these words occur in similar contexts. This is illustrated by the cosine similarities generated from sentiment words in Figures \ref{simliar context} and \ref{Different context}, which is due to the context-dependent nature of pre-trained language models.

\noindent\textbf{Limited Mutual Enhancement of Sentiment Features:} Pre-trained word embeddings exhibit restricted capabilities for mutual enhancement of sentiment-related features, as demonstrated in Figure \ref{Different context}. Despite the contextual presence of some sentiment-related words, the majority of contexts are keyword-irrelevant. These context-based embeddings do not effectively amplify discriminative features when aiding mutual sentiment polarity.

\noindent\textbf{Deficiency of Discriminative Features in Inherent Knowledge:} Due to their pre-training focus on context-based learning, BERT embeddings lack discriminative features for separating sentiment polarities among sentiment-related words as shown in Figure \ref{No context}. Consequently, in short sentences with limited contextual information, these sentiment-discriminative features are not sufficiently pronounced for effective feature learning.

\begin{figure*}[!t] 
\centering
\includegraphics[scale=0.45]{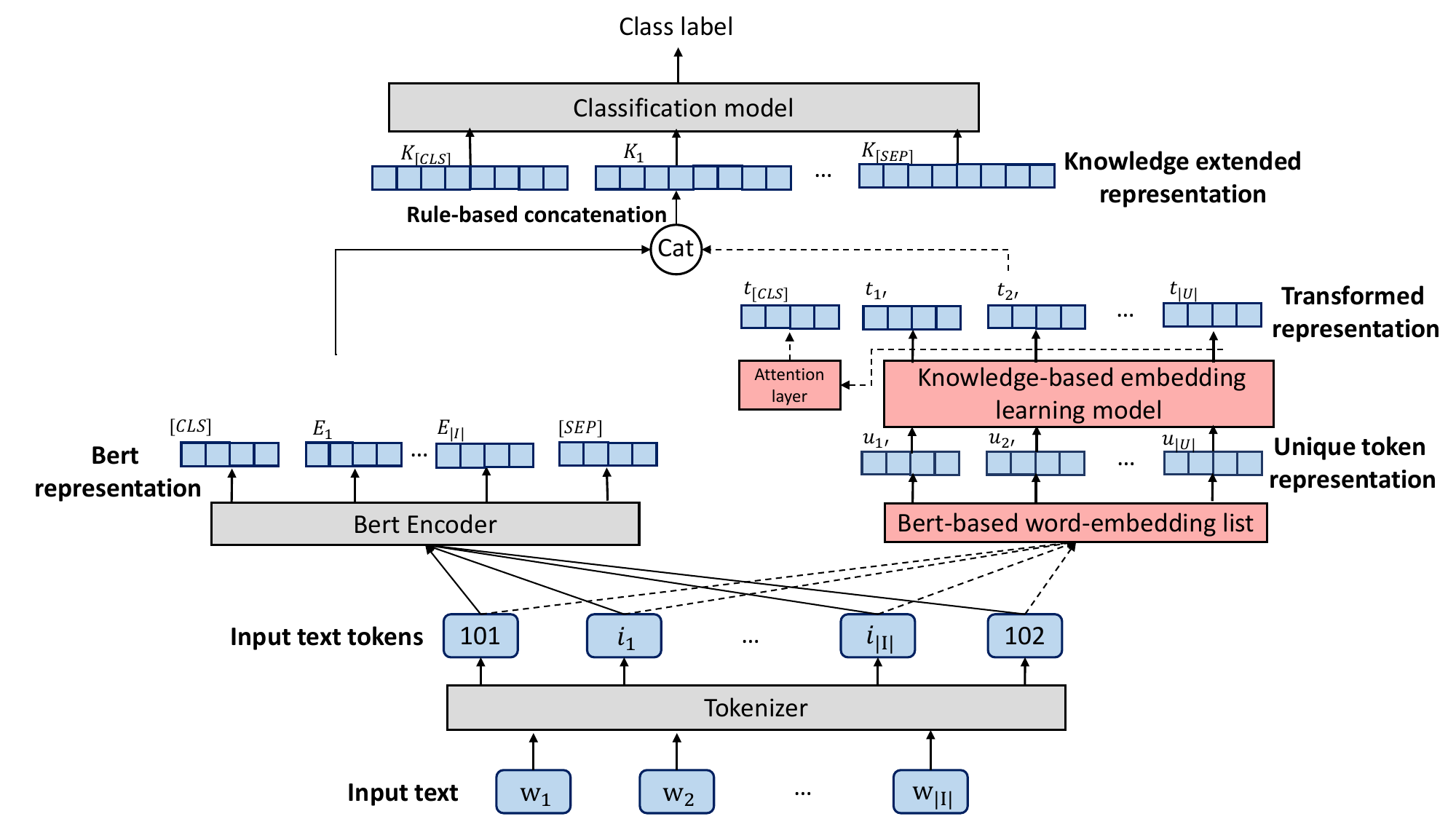}
\caption{Overall architecture of the lexical knowledge-based word embedding learning approach.}
\label{The model architecture of the proposed method.}
\end{figure*}

The limited class-discriminative information in BERT word embeddings can hinder the effectiveness of text classification tasks such as emotion recognition and sentiment analysis, where the ability to distinguish between classes (sentiment features) is crucial \citep{uymaz2022vector,naderalvojoud2020sentiment}. To overcome this limitation, we propose that word embeddings should comprise two components: one capturing contextual information and the other demonstrating class-discriminative power, which is less reliant on context but crucial for enhancing feature learning.

\subsection{Domain Lexical Knowledge-based Word Embedding Learning}
The overall architecture of our proposed method is shown in Figure \ref{The model architecture of the proposed method.}. We develop an embedding learning model, which projects the BERT embedding into a discriminative space, aiming to produce embedding with maximum within-class similarity and between-class separation. The embedding learning model is trained under the domain-specific lexical knowledge collected by our proposed knowledge acquisition algorithm.

\subsubsection{Knowledge Acquisition Algorithm}
In this paper, lexical knowledge refers to the list of words that are closely related to class labels. For example, in sentiment analysis, lexical knowledge is the list of sentiment words expressing positive/negative sentiment. The knowledge acquisition algorithm, detailed in Algorithm \ref{alg:algorithm1}, employs dual word-searching techniques to gather this lexical knowledge from open resources for embedding learning. Initially, it utilizes the related word-searching method\footnote{https://relatedwords.org/} that crawls words containing meaningful relationships with keywords through ConceptNet \citep{speer2017conceptnet} (Algorithm \ref{alg:algorithm1}, lines 1-8). To further enrich the lexical knowledge, the synonym-searching tool\footnote{https://github.com/makcedward/nlpaug} is used to retrieve synonyms from WordNet \citep{pedersen2004wordnet} (Algorithm \ref{alg:algorithm1}, lines 9-21). The output of the related word searching method includes a related score $s_{i}$. We define the input word as the parent of the output word in the two methods. The process begins with label words as initial inputs, expanding the label word space through the related word search to capture diverse perspectives then supplementing them with synonym-searching.

\begin{algorithm}[t]
  \small
  \SetKwData{Left}{left}\SetKwData{This}{this}\SetKwData{Up}{up}
  \SetKwFunction{Union}{Union}\SetKwFunction{FindCompress}{FindCompress}
  \SetKwInOut{Input}{Input}\SetKwInOut{Output}{Output}
  \SetKwInOut{Return}{return}
  \Input{Keywords in the classification tasks (e.g. labels)}
  \Output{Knowledge base $KV$}
  \BlankLine
  Find the related word $w_{i}$ via related word-searching website and the related score $s_{i}$\;
   \While{$w_{i}$ is not a keyword and related score $s_{i} >0$}{
    \uIf{$w_{i}$ is not in the $KV$}
    {   Add $w_{i}$ into the knowledge base $KV$ and labelling it as its parent's label. Save the $w_{i}$-$s_{i}$-label triplet\;}
     \Else 
    {Compare $s_{i}$ with the related score of$w_{i}$ in the triplet, and relabel $w_{i}$ as the label with the higher score. Update the $w_{i}$-$s_{i}$-label triplet if necessary\;}
}
  Find the synonyms $syn_{i}$ of the word $w_{i}$ via the synonym searching tool. Create an empty list $L$\;
  \uIf{$syn_{i}$ is not in the $KV$ and $L$ }
  {
    Add $syn_{i}$ to $KV$ and label it with its parent’s label. Also, add $syn_{i}$ to $L$\;
    }
    \uElseIf{$syn_{i}$ in the $KV$ and $L$ }
      {
       \uIf{The parent of $syn_{i}$ has the same label as $syn_{i}$ had in the $KV$}
             {
             Pass, as the word $syn_{i}$ is repeated in $KV$\; 
             }
             \uElse 
             {
              Delete it from $KV$ as it is a confusing word\; 
             }
        }
   \uElseIf{$syn_{i}$ is not in the $KV$ but is in $L$ }
    {         
        Pass, as it is a confusing word and has been deleted from $KV$\;      
    } 

    \Else 
    {
      Add $syn_{i}$ in $L$\;
   }
   Remove the words in $KV$ that contain sub-pieces in BERT's representation\;
  \Return{$KV$}
  \caption{Knowledge acquisition algorithm}
 \label{alg:algorithm1}
\end{algorithm}\DecMargin{1em}
\normalsize

\subsubsection{BERT Word Embedding Pre-processing\label{sec:BERT word embedding pre-processing}} 
BERT's vocabulary contains 30522 tokens, including words and sub-pieces (segmented words with the prefix `\#\#'). We denote the word without `\#\#' as a unique word and remove the unused slots, mask tokens, sub-piece, and meaningless tokens. Finally, 21764 unique tokens are obtained. Except for the words acquired by the knowledge acquisition algorithm for each class, the other unique tokens are added to the lexicon and labeled as `neutral'.
To attain a fixed representation for general use, we build a BERT-based word-embedding list. 
We input the unique tokens into the BERT model separately, and the output vectors obtained include the embedding of $ \left [CLS  \right ]$, [token's index], and $ \left [SEP  \right ]$. Here, \( \left[ \text{CLS} \right] \) is a classification token added at the beginning of the input sequence to capture overall sentence-level information, and \( \left[ \text{SEP} \right] \) is a separator token used to mark the end of a sequence or separate two sequences in paired inputs. Both tokens are pre-defined in BERT and serve specific roles in its architecture. However, as our goal is to obtain a representation of the word itself rather than sentence-level or sequence-ending information, we choose the vector of the index as the token's word representation. We denote it as word-unique embedding in the following sub-sections. The index and the embedding are saved in a word-embedding list.

\subsubsection{Knowledge-based Word Embedding Learning} 
The knowledge-based word embedding learning is based on a five-layer neural network, where the output of the second layer in the embedding learning model is adopted as word representation, while the output of the final layer is the class label.
We therefore employ two loss functions: the center loss applied to the output of the second layer and the cross-entropy loss applied to the output of the final layer. The distance function in the center loss can be either Euclidean Distance or Cosine Similarity. The detailed network information is summarized in Table \ref{The multi-layer neutral network details.}.

\begin{table}[htbp]
\centering
\caption{Settings of the knowledge-based word embedding learning network, where $|Class|$ is the number of classes.}
\normalsize
\resizebox{\linewidth}{!}{
\begin{tabular}{ccccc}
\hline
Layer        & Input dimension & Output dimension   & Activation function & Loss function     \\ \hline
Linear layer & 768             & 512                & ReLU                &                   \\
Linear layer & 512             & 768                & ReLU                & Center loss       \\
Linear layer & 768             & 512                & ReLU                &                   \\
Linear layer & 512             & 300                & ReLU                &                   \\
Linear layer & 300             & $|Class|$ & Sigmoid             & Cross-entropy loss \\ \hline
\end{tabular}}
\label{The multi-layer neutral network details.}
\end{table}

The knowledge-based word embedding learning model projects the BERT word embedding to a more discriminative space based on the available lexical knowledge. In this space, the similarity of words' embeddings within the same class is maximized, while the similarity of words between different classes is minimized via the center loss functions. For neutral words belonging to none of the domain-specific classes, we do not attempt to project them into a single cluster because they are semantically very different. 

We therefore formulate the center loss function as Eqns \eqref{eq:centerlossmdistance} and \eqref{eq:centerlossmcossim} for Euclidean Distance measure and Cosine Similarity measure in the new space, respectively:

\begin{equation}
 Loss_{Dist}=\sum_{q=1}^{L}\sum_{x_{k}\in l_{q}}^{}(1-y_{k})(x_{k}-c_{q})^{2}
	 \label{eq:centerlossmdistance}
	\end{equation} 	
	
\begin{equation}
\begin{aligned}
\mathrm{Loss}_{\mathrm{Cosine}} = \sum_{q=1}^{L} \sum_{x_k \in l_q} (1 - y_k)(1 - \cos(x_k, c_q))
\end{aligned}
\label{eq:centerlossmcossim}
\end{equation}
	
where $L$ is the number of classes, $l_{q}$ denotes the class label of $q$-th class, and $c_{q}$ denotes the mean embedding of the lexicon of class $l_{q}$. $x_{k}$ is the embedding of the $k$-th input word belonging to class $l_{q}$. $y_{k}=1$ if the $l_{q}$ is `neutral', otherwise,  $y_{k}=0$.

We consider both word-level and sentence-level applications of our embedding learning model. For word-level application, the input sentence is tokenized as $I=\left [i_{1}, i_{2}, \cdots , i_{\left| I \right|}\right ]$. If $i_{j}$ is in the word-embedding list, its corresponding word-unique embedding $u_{j}$ is modified by the knowledge-based embedding learning model to obtain a new embedding $t_{j}$. The word-embedding list and word-unique embedding are obtained from BERT word embedding pre-processing. The knowledge-based word representation for the input sentence is obtained as $T=\left [t_{1^{'}}, t_{2^{'}}, \cdots , t_{\left| U \right|}\right ]$, where $\left| U \right|$ $\left ( \left| U \right|\leq \left|I \right| \right )$ is the number of the words of the input text appearing in the word-embedding list.

For the sentence-level application, we apply an attention layer on the $T$ to obtain knowledge-based sentence representation $t_{CLS}$. The attention layer is trainable in the classification model, while the knowledge-based learning model is fixed once learned.

To deploy the new representation into text classification, the rules are as follows to build enhanced embeddings with both contextual and domain-knowledge information:

\noindent 1. In word-level usage, concatenating the BERT pre-trained word embedding $E$ with the knowledge-based embedding $T$. If the word does not exist in the word-embedding list, self-concatenating of its BERT embedding is performed.

\noindent 2. In sentence-level usage, concatenating the BERT pre-trained sentence embedding $\left [CLS  \right ]$ with $t_{CLS}$.

\section{Experiments}

We assess our approach through three distinct classification tasks: sentiment analysis, emotion recognition, and question answering. In this section, we begin by examining the efficacy of the word embedding learning model using similarity measurements. Subsequently, we integrate the learning model into various classification models to evaluate their performance in classification tasks. We refer to our enhanced BERT-based Contextual-Knowledgeable Embedding as BERTCK in the following sections.

\subsection{Evaluation of Word Embedding Learning Efficacy}

In the word embedding learning model, the sentiment lexicon is obtained from \cite{liu2005opinion}, while excluding words represented as sub-pieces in BERT. The lexicons for emotion recognition and question answering are collected using our knowledge acquisition algorithm. The labels of the lexicon are based on the evaluated datasets. Although question answering does not have a `neutral' class, all the unique tokens, not captured by the acquisition algorithm are considered as `neutral' class. This ensures comprehensive BERT token usage for lexicon-based embedding learning. The details are provided in Table \ref{Information of domain-specific lexicons}. The proposed knowledge-based embedding learning model is trained with a dropout rate of 0.4 and a learning rate of 5e-5.

\begin{table*}[htbp]
\centering
\caption{Information of domain-specific lexicons for sentiment analysis, emotion recognition, and question answering.}
\normalsize
\resizebox{\linewidth}{!}{
\begin{tabular}{cccccccccc}
\hline
            & \multicolumn{9}{c}{Sentiment Analysis}                                                        \\ \cline{2-10} 
Labels      &       &          &       & positive    & negative     & neutral &        &          &         \\ \cline{2-10} 
Lexicon size &       &          &       & 900         & 1518         & 19346   &        &          &         \\ \hline
            & \multicolumn{9}{c}{Emotion Recognition}                                                       \\ \hline
Labels      & anger & sad      & fear  & happy       & boredom      & worry   & love   & surprise & neutral \\ \cline{2-10} 
Lexicon size & 447   & 368      & 277   & 718         & 17           & 104     & 173    & 64       & 19596   \\ \hline
            & \multicolumn{9}{c}{Questing Answering}                                                        \\ \hline
Labels      &       & location & human & description & abbreviation & numeric & entity & neutral  &         \\ \cline{2-10} 
Lexicon size &       & 236      & 199   & 172         & 12           & 431     & 1066   & 19648    &         \\ \hline
\end{tabular}}
\label{Information of domain-specific lexicons}
\end{table*}

\subsubsection{Similarity Measures and Analysis}
The knowledge-based embedding learning model aims to maximize within-class similarity and minimize between-class similarity using available lexical knowledge. The model is evaluated based on the changes in within-class and between-class embedding similarity before and after knowledge-based learning. For Cosine Similarity-based evaluation, increased within-class and decreased between-class measures mean improvement. For Euclidean Distance-based evaluation, decreased within-class and increased between-class measures mean improvement.

Table \ref{mappingresultsa}, \ref{mappingresultER} and \ref{mappingresultTREC} show the evaluation results for the sentiment analysis lexicon, emotion recognition lexicon, and the question answering (TREC) lexicon, respectively.

\begin{table}[htbp]
\centering
\caption{Within-class and between-class similarity measures of knowledge-based word embedding of sentimental lexicon}
\resizebox{\linewidth}{!}{
\begin{tabular}{cc|ccc}
\hline
                                               &        & positive                                                    & negative                                                     & neutral                                                     \\ \hline
\multicolumn{1}{c|}{\multirow{2}{*}{positive}} & Dist   & \begin{tabular}[c]{@{}c@{}}0.9302\\ (-10.3896)\end{tabular} & \begin{tabular}[c]{@{}c@{}}22.5643\\ (+11.3057)\end{tabular} & \begin{tabular}[c]{@{}c@{}}13.7138\\ (+1.4725)\end{tabular} \\ \cline{2-5} 
\multicolumn{1}{c|}{}                          & Cosine & \begin{tabular}[c]{@{}c@{}}0.9387\\ (+0.2719)\end{tabular}  & \begin{tabular}[c]{@{}c@{}}0.3337\\ (-0.3348)\end{tabular}   & \begin{tabular}[c]{@{}c@{}}0.5921\\ (-0.0306)\end{tabular}  \\ \hline
\multicolumn{1}{c|}{\multirow{2}{*}{negative}} & Dist   &                                                             & \begin{tabular}[c]{@{}c@{}}0.7504\\ (-10.1221)\end{tabular}  & \begin{tabular}[c]{@{}c@{}}12.1239\\ (+0.1093)\end{tabular} \\ \cline{2-5} 
\multicolumn{1}{c|}{}                          & Cosine &                                                             & \begin{tabular}[c]{@{}c@{}}0.9496\\ (+0.2615)\end{tabular}   & \begin{tabular}[c]{@{}c@{}}0.6194\\ (-0.0143)\end{tabular}  \\ \hline
\multicolumn{1}{c|}{\multirow{2}{*}{nertral}}  & Dist   &                                                             &                                                              & \begin{tabular}[c]{@{}c@{}}8.9415\\ (-3.8194)\end{tabular}  \\ \cline{2-5} 
\multicolumn{1}{c|}{}                          & Cosine &                                                             &                                                              & \begin{tabular}[c]{@{}c@{}}0.7649\\ (+0.1634)\end{tabular}  \\ \hline
\end{tabular}
}

\label{mappingresultsa}
\end{table}

\begin{table*}[h]
\centering

\caption{Within-class and between-class similarity measures of embedding of emotional lexicon}
\resizebox{\linewidth}{!}{
\begin{tabular}{lc|ccccccccc}
\hline
\multicolumn{1}{c}{}                           & \multicolumn{1}{l|}{} & Anger                                                      & Fear                                                         & Sadness                                                      & Joy                                                          & Love                                                         & Worry                                                        & Boredom                                                      & Surprise                                                     & Neutral                                                     \\ \hline
\multicolumn{1}{l|}{\multirow{2}{*}{Anger}}    & Dist                  & \begin{tabular}[c]{@{}c@{}}0.9448\\ (-9.7420)\end{tabular} & \begin{tabular}[c]{@{}c@{}}22.2888\\ (+11.6290)\end{tabular} & \begin{tabular}[c]{@{}c@{}}23.1294\\ (+12.0888)\end{tabular} & \begin{tabular}[c]{@{}c@{}}23.1089\\ (+11.9000)\end{tabular} & \begin{tabular}[c]{@{}c@{}}23.1065\\ (+12.1600)\end{tabular} & \begin{tabular}[c]{@{}c@{}}22.3857\\ (+12.2430)\end{tabular} & \begin{tabular}[c]{@{}c@{}}22.2377\\ (+11.8931)\end{tabular} & \begin{tabular}[c]{@{}c@{}}21.0970\\ (+10.5667)\end{tabular} & \begin{tabular}[c]{@{}c@{}}17.7749\\ (+5.7117)\end{tabular} \\ \cline{2-11} 
\multicolumn{1}{l|}{}                          & Cosine                 & \begin{tabular}[c]{@{}c@{}}0.9104\\ (+0.2100)\end{tabular} & \begin{tabular}[c]{@{}c@{}}0.3651\\ (-0.3370)\end{tabular}   & \begin{tabular}[c]{@{}c@{}}0.3070\\ (-0.3764)\end{tabular}   & \begin{tabular}[c]{@{}c@{}}0.2869\\ (-0.3900)\end{tabular}   & \begin{tabular}[c]{@{}c@{}}0.2823\\ (-0.4100)\end{tabular}   & \begin{tabular}[c]{@{}c@{}}0.3486\\ (-0.3780)\end{tabular}   & \begin{tabular}[c]{@{}c@{}}0.3171\\ (-0.4477)\end{tabular}   & \begin{tabular}[c]{@{}c@{}}0.3783\\ (-0.3325)\end{tabular}   & \begin{tabular}[c]{@{}c@{}}0.3914\\ (-0.2395)\end{tabular}  \\ \hline
\multicolumn{1}{l|}{\multirow{2}{*}{Fear}}     & Dist                  &                                                            & \begin{tabular}[c]{@{}c@{}}1.0431\\ (-9.4290)\end{tabular}   & \begin{tabular}[c]{@{}c@{}}22.3475\\ (+11.4543)\end{tabular} & \begin{tabular}[c]{@{}c@{}}22.3354\\ (+11.3000)\end{tabular} & \begin{tabular}[c]{@{}c@{}}22.8561\\ (+12.0400)\end{tabular} & \begin{tabular}[c]{@{}c@{}}21.8859\\ (+11.8990)\end{tabular} & \begin{tabular}[c]{@{}c@{}}20.9833\\ (+10.7931)\end{tabular} & \begin{tabular}[c]{@{}c@{}}21.2005\\ (+10.8227)\end{tabular} & \begin{tabular}[c]{@{}c@{}}18.3406\\ (+6.3999)\end{tabular} \\ \cline{2-11} 
\multicolumn{1}{l|}{}                          & Cosine                 & \multicolumn{1}{l}{}                                       & \begin{tabular}[c]{@{}c@{}}0.8803\\ (+0.1672)\end{tabular}   & \begin{tabular}[c]{@{}c@{}}0.3778\\ (-0.3137)\end{tabular}   & \begin{tabular}[c]{@{}c@{}}0.2684\\ (-0.4100)\end{tabular}   & \begin{tabular}[c]{@{}c@{}}0.2634\\ (-0.4360)\end{tabular}   & \begin{tabular}[c]{@{}c@{}}0.3959\\ (-0.3990)\end{tabular}   & \begin{tabular}[c]{@{}c@{}}0.3204\\ (-0.4535)\end{tabular}   & \begin{tabular}[c]{@{}c@{}}0.3860\\ (-0.3329)\end{tabular}   & \begin{tabular}[c]{@{}c@{}}0.3479\\ (-0.2897)\end{tabular}  \\ \hline
\multicolumn{1}{l|}{\multirow{2}{*}{Sadness}}  & Dist                  &                                                            &                                                              & \begin{tabular}[c]{@{}c@{}}1.0028\\ (-10.1240)\end{tabular}  & \begin{tabular}[c]{@{}c@{}}22.9890\\ (+11.6000)\end{tabular} & \begin{tabular}[c]{@{}c@{}}22.9962\\ (+11.8700)\end{tabular} & \begin{tabular}[c]{@{}c@{}}21.7637\\ (+11.3810)\end{tabular} & \begin{tabular}[c]{@{}c@{}}22.1213\\ (+10.6566)\end{tabular} & \begin{tabular}[c]{@{}c@{}}21.9769\\ (+11.1025)\end{tabular} & \begin{tabular}[c]{@{}c@{}}17.4519\\ (+5.2372)\end{tabular} \\ \cline{2-11} 
\multicolumn{1}{l|}{}                          & Cosine                 & \multicolumn{1}{l}{}                                       & \multicolumn{1}{l}{}                                         & \begin{tabular}[c]{@{}c@{}}0.9086\\ (+0.2257)\end{tabular}   & \begin{tabular}[c]{@{}c@{}}0.3077\\ (-0.3600)\end{tabular}   & \begin{tabular}[c]{@{}c@{}}0.3473\\ (-0.3380)\end{tabular}   & \begin{tabular}[c]{@{}c@{}}0.4124\\ (-0.3040)\end{tabular}   & \begin{tabular}[c]{@{}c@{}}0.7105\\ (-0.0505)\end{tabular}   & \begin{tabular}[c]{@{}c@{}}0.2676\\ (-0.4267)\end{tabular}   & \begin{tabular}[c]{@{}c@{}}0.4382\\ (-0.1875)\end{tabular}  \\ \hline
\multicolumn{1}{l|}{\multirow{2}{*}{Joy}}      & Dist                  &                                                            &                                                              &                                                              & \begin{tabular}[c]{@{}c@{}}0.8835\\ (+10.4000)\end{tabular}  & \begin{tabular}[c]{@{}c@{}}24.2756\\ (+13.1800)\end{tabular} & \begin{tabular}[c]{@{}c@{}}22.9640\\ (+12.4730)\end{tabular} & \begin{tabular}[c]{@{}c@{}}21.7347\\ (+11.1066)\end{tabular} & \begin{tabular}[c]{@{}c@{}}22.5065\\ (+11.7061)\end{tabular} & \begin{tabular}[c]{@{}c@{}}12.1184\\ (-0.0950)\end{tabular} \\ \cline{2-11} 
\multicolumn{1}{l|}{}                          & Cosine                 & \multicolumn{1}{l}{}                                       & \multicolumn{1}{l}{}                                         & \multicolumn{1}{l}{}                                         & \begin{tabular}[c]{@{}c@{}}0.9341\\ (+0.2600)\end{tabular}   & \begin{tabular}[c]{@{}c@{}}0.3348\\ (-0.3530)\end{tabular}   & \begin{tabular}[c]{@{}c@{}}0.2457\\ (-0.4650)\end{tabular}   & \begin{tabular}[c]{@{}c@{}}0.4433\\ (-0.3087)\end{tabular}   & \begin{tabular}[c]{@{}c@{}}0.3643\\ (-0.3349)\end{tabular}   & \begin{tabular}[c]{@{}c@{}}0.6297\\ (+0.0037)\end{tabular}  \\ \hline
\multicolumn{1}{l|}{\multirow{2}{*}{Love}}     & Dist                  &                                                            &                                                              &                                                              &                                                              & \begin{tabular}[c]{@{}c@{}}1.1314\\ (-9.5370)\end{tabular}   & \begin{tabular}[c]{@{}c@{}}21.9449\\ (+11.7430)\end{tabular} & \begin{tabular}[c]{@{}c@{}}22.6103\\ (+12.3039)\end{tabular} & \begin{tabular}[c]{@{}c@{}}21.8413\\ (+11.2833)\end{tabular} & \begin{tabular}[c]{@{}c@{}}19.9816\\ (+7.8818)\end{tabular} \\ \cline{2-11} 
\multicolumn{1}{l|}{}                          & Cosine                 & \multicolumn{1}{l}{}                                       & \multicolumn{1}{l}{}                                         & \multicolumn{1}{l}{}                                         & \multicolumn{1}{l}{}                                         & \begin{tabular}[c]{@{}c@{}}0.8899\\ (+0.1760)\end{tabular}   & \begin{tabular}[c]{@{}c@{}}0.3930\\ (-0.3350)\end{tabular}   & \begin{tabular}[c]{@{}c@{}}0.5241\\ (-0.2475)\end{tabular}   & \begin{tabular}[c]{@{}c@{}}0.3757\\ (-0.3384)\end{tabular}   & \begin{tabular}[c]{@{}c@{}}0.3589\\ (-0.2744)\end{tabular}  \\ \hline
\multicolumn{1}{l|}{\multirow{2}{*}{Worry}}    & Dist                  &                                                            &                                                              &                                                              &                                                              &                                                              & \begin{tabular}[c]{@{}c@{}}1.1423\\ (-8.0850)\end{tabular}   & \begin{tabular}[c]{@{}c@{}}21.6753\\ (+12.1748)\end{tabular} & \begin{tabular}[c]{@{}c@{}}20.9050\\ (+11.2127)\end{tabular} & \begin{tabular}[c]{@{}c@{}}18.8816\\ (+7.4277)\end{tabular} \\ \cline{2-11} 
\multicolumn{1}{l|}{}                          & Cosine                 & \multicolumn{1}{l}{}                                       & \multicolumn{1}{l}{}                                         & \multicolumn{1}{l}{}                                         & \multicolumn{1}{l}{}                                         & \multicolumn{1}{l}{}                                         & \begin{tabular}[c]{@{}c@{}}0.8359\\ (+0.0662)\end{tabular}   & \begin{tabular}[c]{@{}c@{}}0.5756\\ (-0.2294)\end{tabular}   & \begin{tabular}[c]{@{}c@{}}0.4574\\ (-0.2907)\end{tabular}   & \begin{tabular}[c]{@{}c@{}}0.4929\\ (-0.1681)\end{tabular}  \\ \hline
\multicolumn{1}{l|}{\multirow{2}{*}{Boredom}}  & Dist                  &                                                            &                                                              &                                                              &                                                              &                                                              &                                                              & \begin{tabular}[c]{@{}c@{}}1.3701\\ (-8.0716)\end{tabular}   & \begin{tabular}[c]{@{}c@{}}20.7863\\ (+10.7682)\end{tabular} & \begin{tabular}[c]{@{}c@{}}18.6455\\ (+7.0605)\end{tabular} \\ \cline{2-11} 
\multicolumn{1}{l|}{}                          & Cosine                 & \multicolumn{1}{l}{}                                       & \multicolumn{1}{l}{}                                         & \multicolumn{1}{l}{}                                         & \multicolumn{1}{l}{}                                         & \multicolumn{1}{l}{}                                         & \multicolumn{1}{l}{}                                         & \begin{tabular}[c]{@{}c@{}}0.8421\\ (+0.0089)\end{tabular}   & \begin{tabular}[c]{@{}c@{}}0.3479\\ (-0.3955)\end{tabular}   & \begin{tabular}[c]{@{}c@{}}0.3492\\ (-0.3077)\end{tabular}  \\ \hline
\multicolumn{1}{l|}{\multirow{2}{*}{Surprise}} & Dist                  &                                                            &                                                              &                                                              &                                                              &                                                              &                                                              &                                                              & \begin{tabular}[c]{@{}c@{}}1.3642\\ (-8.2168)\end{tabular}   & \begin{tabular}[c]{@{}c@{}}18.5990\\ (+6.7635)\end{tabular} \\ \cline{2-11} 
\multicolumn{1}{l|}{}                          & Cosine                 & \multicolumn{1}{l}{}                                       & \multicolumn{1}{l}{}                                         & \multicolumn{1}{l}{}                                         & \multicolumn{1}{l}{}                                         & \multicolumn{1}{l}{}                                         & \multicolumn{1}{l}{}                                         & \multicolumn{1}{l}{}                                         & \begin{tabular}[c]{@{}c@{}}0.8182\\ (+0.0592)\end{tabular}   & \begin{tabular}[c]{@{}c@{}}0.3415\\ (-0.2943)\end{tabular}  \\ \hline
\multicolumn{1}{l|}{\multirow{2}{*}{Neutral}}  & Dist                  & \multicolumn{1}{l}{}                                       & \multicolumn{1}{l}{}                                         & \multicolumn{1}{l}{}                                         & \multicolumn{1}{l}{}                                         & \multicolumn{1}{l}{}                                         & \multicolumn{1}{l}{}                                         & \multicolumn{1}{l}{}                                         & \multicolumn{1}{l}{}                                         & \begin{tabular}[c]{@{}c@{}}11.3803\\ (-2.2828)\end{tabular} \\ \cline{2-11} 
\multicolumn{1}{l|}{}                          & Cosine                 &                                                            &                                                              &                                                              &                                                              &                                                              &                                                              &                                                              &                                                              & \begin{tabular}[c]{@{}c@{}}0.7099\\ (+0.1581)\end{tabular}  \\ \hline
\end{tabular}

}

\label{mappingresultER}
\end{table*}

\begin{table*}[h]
\centering
\caption{Within-class and between-class similarity measures of embedding of question answering (TREC) lexicon}
\resizebox{\linewidth}{!}
{\begin{tabular}{lc|ccccccc}
\hline
\multicolumn{1}{c}{}                               &       & abbreviation                                               & entity                                                      & destination                                                 & human                                                       & location                                                    & numeric                                                     & neutral                                                     \\ \hline
\multicolumn{1}{l|}{\multirow{2}{*}{abbreviation}} & Dist  & \begin{tabular}[c]{@{}c@{}}9.4561\\ (-3.2310)\end{tabular} & \begin{tabular}[c]{@{}c@{}}14.7852\\ (+2.1032)\end{tabular} & \begin{tabular}[c]{@{}c@{}}12.2542\\ (-0.0198)\end{tabular} & \begin{tabular}[c]{@{}c@{}}12.4177\\ (-0.4766)\end{tabular} & \begin{tabular}[c]{@{}c@{}}13.7485\\ (+0.2969)\end{tabular} & \begin{tabular}[c]{@{}c@{}}14.6744\\ (+1.8996)\end{tabular} & \begin{tabular}[c]{@{}c@{}}12.1724\\ (-0.5313)\end{tabular} \\ \cline{2-9} 
\multicolumn{1}{l|}{}                              & Cosine & \begin{tabular}[c]{@{}c@{}}0.7958\\ (+0.0881)\end{tabular} & \begin{tabular}[c]{@{}c@{}}0.5229\\ (-0.0838)\end{tabular}  & \begin{tabular}[c]{@{}c@{}}0.5675\\ (-0.0621)\end{tabular}  & \begin{tabular}[c]{@{}c@{}}0.5384\\ (-0.0589)\end{tabular}  & \begin{tabular}[c]{@{}c@{}}0.5186\\ (-0.0453)\end{tabular}  & \begin{tabular}[c]{@{}c@{}}0.4680\\ (-0.1348)\end{tabular}  & \begin{tabular}[c]{@{}c@{}}0.7051\\ (+0.0952)\end{tabular}  \\ \hline
\multicolumn{1}{l|}{\multirow{2}{*}{entity}}       & Dist  &                                                            & \begin{tabular}[c]{@{}c@{}}4.0511\\ (-8.5271)\end{tabular}  & \begin{tabular}[c]{@{}c@{}}15.9270\\ (+3.6725)\end{tabular} & \begin{tabular}[c]{@{}c@{}}15.3216\\ (+2.5178)\end{tabular} & \begin{tabular}[c]{@{}c@{}}16.6753\\ (+3.4037)\end{tabular} & \begin{tabular}[c]{@{}c@{}}16.9908\\ (+4.2633)\end{tabular} & \begin{tabular}[c]{@{}c@{}}20.0775\\ (+7.4325)\end{tabular} \\ \cline{2-9} 
\multicolumn{1}{l|}{}                              & Cosine &                                                            & \begin{tabular}[c]{@{}c@{}}0.9697\\ (+0.3671)\end{tabular}  & \begin{tabular}[c]{@{}c@{}}0.4774\\ (-0.1416)\end{tabular}  & \begin{tabular}[c]{@{}c@{}}0.5012\\ (-0.0905)\end{tabular}  & \begin{tabular}[c]{@{}c@{}}0.4653\\ (-0.0984)\end{tabular}  & \begin{tabular}[c]{@{}c@{}}0.4471\\ (-0.1467)\end{tabular}  & \begin{tabular}[c]{@{}c@{}}0.3456\\ (-0.2566)\end{tabular}  \\ \hline
\multicolumn{1}{l|}{\multirow{2}{*}{destination}}  & Dist  &                                                            &                                                             & \begin{tabular}[c]{@{}c@{}}8.4670\\ (-3.2577)\end{tabular}  & \begin{tabular}[c]{@{}c@{}}12.7209\\ (+0.2800)\end{tabular} & \begin{tabular}[c]{@{}c@{}}14.9462\\ (+1.8923)\end{tabular} & \begin{tabular}[c]{@{}c@{}}14.3753\\ (+2.0632)\end{tabular} & \begin{tabular}[c]{@{}c@{}}15.3301\\ (+3.0651)\end{tabular} \\ \cline{2-9} 
\multicolumn{1}{l|}{}                              & Cosine &                                                            &                                                             & \begin{tabular}[c]{@{}c@{}}0.8099\\ (+0.1549)\end{tabular}  & \begin{tabular}[c]{@{}c@{}}0.5675\\ (-0.0475)\end{tabular}  & \begin{tabular}[c]{@{}c@{}}0.4844\\ (-0.0899)\end{tabular}  & \begin{tabular}[c]{@{}c@{}}0.5310\\ (-0.0862)\end{tabular}  & \begin{tabular}[c]{@{}c@{}}0.5850\\ (-0.0382)\end{tabular}  \\ \hline
\multicolumn{1}{l|}{\multirow{2}{*}{human}}        & Dist  &                                                            &                                                             &                                                             & \begin{tabular}[c]{@{}c@{}}7.0123\\ (-5.9012)\end{tabular}  & \begin{tabular}[c]{@{}c@{}}14.8093\\ (+1.3484)\end{tabular} & \begin{tabular}[c]{@{}c@{}}14.6985\\ (+2.0870)\end{tabular} & \begin{tabular}[c]{@{}c@{}}16.1124\\ (+3.2529)\end{tabular} \\ \cline{2-9} 
\multicolumn{1}{l|}{}                              & Cosine &                                                            &                                                             &                                                             & \begin{tabular}[c]{@{}c@{}}0.8527\\ (+0.2588)\end{tabular}  & \begin{tabular}[c]{@{}c@{}}0.4762\\ (-0.0794)\end{tabular}  & \begin{tabular}[c]{@{}c@{}}0.4896\\ (-0.0967)\end{tabular}  & \begin{tabular}[c]{@{}c@{}}0.5192\\ (-0.0737)\end{tabular}  \\ \hline
\multicolumn{1}{l|}{\multirow{2}{*}{location}}     & Dist  &                                                            &                                                             &                                                             &                                                             & \begin{tabular}[c]{@{}c@{}}7.2540\\ (-6.3636)\end{tabular}  & \begin{tabular}[c]{@{}c@{}}16.0882\\ (+2.6716)\end{tabular} & \begin{tabular}[c]{@{}c@{}}17.6993\\ (+4.3762)\end{tabular} \\ \cline{2-9} 
\multicolumn{1}{l|}{}                              & Cosine &                                                            &                                                             &                                                             &                                                             & \begin{tabular}[c]{@{}c@{}}0.8571\\ (+0.3063)\end{tabular}  & \begin{tabular}[c]{@{}c@{}}0.4583\\ (-0.0973)\end{tabular}  & \begin{tabular}[c]{@{}c@{}}0.4612\\ (-0.1040)\end{tabular}  \\ \hline
\multicolumn{1}{l|}{\multirow{2}{*}{numeric}}      & Dist  &                                                            &                                                             &                                                             &                                                             &                                                             & \begin{tabular}[c]{@{}c@{}}6.3676\\ (-6.3716)\end{tabular}  & \begin{tabular}[c]{@{}c@{}}18.392\\ (+5.6558)\end{tabular}  \\ \cline{2-9} 
\multicolumn{1}{l|}{}                              & Cosine &                                                            &                                                             &                                                             &                                                             &                                                             & \begin{tabular}[c]{@{}c@{}}0.9020\\ (+0.3050)\end{tabular}  & \begin{tabular}[c]{@{}c@{}}0.4187\\ (-0.1795)\end{tabular}  \\ \hline
\multicolumn{1}{l|}{\multirow{2}{*}{neutral}}      & Dist  &                                                            &                                                             &                                                             &                                                             &                                                             &                                                             & \begin{tabular}[c]{@{}c@{}}1.9505\\ (-10.6496)\end{tabular} \\ \cline{2-9} 
\multicolumn{1}{l|}{}                              & Cosine &                                                            &                                                             &                                                             &                                                             &                                                             &                                                             & \begin{tabular}[c]{@{}c@{}}0.9962\\ (+0.3867)\end{tabular}  \\ \hline
\end{tabular}
}
\label{mappingresultTREC}
\end{table*}

The results of Cosine Similarity (Cosine) are for the model trained by loss function $Loss_{Cosine}$ while results of Euclidean Distance (Dist) are for the model trained by loss function $Loss_{Dist}$. The values inside the brackets indicate the similarity change after knowledge-based embedding learning.

It is observed that the within-class similarity and between-class difference have been significantly improved after knowledge-based embedding learning. This improvement could facilitate the subsequent classification task. Since we do not expect the neutral words to form a cluster in the new discriminative space, the similarity in this group has changed slightly. 

\subsection{Evaluation of Classification Performance}
\subsubsection{Datasets} We assess our approach using six benchmark text classification datasets. For sentiment analysis, we evaluate the three binary datasets including SST-2 \citep{socher2013recursive},  CR \citep{ding2008holistic}, and RT \citep{pang2005seeing}. For emotion recognition, we access two seven-class datasets ISEAR \citep{scherer1994evidence} and AMAN \citep{aman2008using}. For question answering, we evaluate TREC \citep{li2002learning}, a dataset that encompasses questions from six distinct domains. The characteristics of these datasets are detailed in Table \ref{Statistics for the six text classification datasets.}. Our experimental design involves varying the training data volume, where we randomly select 20\%, 40\%, 60\%, and 80\% of the original training sets to discern the efficacy of our method in scenarios with limited training data. In cases where a dataset lacks a predefined validation set, we allocate half of the chosen training data for validation purposes.

\begin{table*}[htbp]
\caption{Statistics for the six text classification datasets.}
\centering
\resizebox{\linewidth}{!}{
\begin{tabular}{cccccc}
\hline
Application                          & Dataset & \#Class                                                                    & \#Train & \#Dev & \#Test \\ \hline
\multirow{3}{*}{Sentiment Analysis}  & SST2    & \multirow{3}{*}{2: positive, and negative}                                 & 7447    & -     & 1821   \\
                                     & CR      &                                                                            & 3394    & -     & 376    \\
                                     & RT      &                                                                            & 8636    & 960   & 1607   \\ \hline
\multirow{2}{*}{Emotion Recognition} & ISEAR   & 7: joy, sadness, fear, anger, guilt, disgust, and shame                    & 6133    & -     & 1533   \\
                                     & AMAN    & 7: angry, disgust, happy, neutral, surprise, sad, and fear                 & 3272    & -     & 818    \\ \hline
Question Answering                   & TREC    & 7: location, entity, numeric,human, description, abbreviation, and neutral & 4907    & 546   & 500    \\ \hline
\end{tabular}
}
\label{Statistics for the six text classification datasets.}
\end{table*}

\subsubsection{Implementation Details}

We evaluate our knowledge-based embedding learning method using four classification models to show the method's generality:

\textbf{BiLstm-att} \citep{lin2017structured}: A BiLSTM model with a unique regularization term and a self-attentive sentence representation learning mechanism for text classification,

\textbf{LCL} \citep{suresh2021not}: A label-aware contrastive learning model for text classification. 

\textbf{DualCL} \citep{chen2022dual}: A refined BiLSTM-CNN dual-channel model that intricately extracts features to optimize cluster relationships for enhanced multi-class text classification. 

\textbf{Kil} \citep{zhang2021improving}: A knowledge-enhanced model that incorporates label-knowledge into classifier by relatedness calculation.

We compare our BERT-based Contextual-Knowledgeable
Embedding (BERTCK) with other word embedding methods including BERT, Roberta \citep{liu2019roberta} and CoSE \citep{wang2022contextual}, a BERT-based contextual sentiment embedding trained on Sentiment140 \citep{go2009twitter} by Bi-GRU (only compared in sentiment analysis).

A hidden dimension of 256 in the BiLstm-att model is adopted with the learning rate set to 9e-4 and the dropout rate set to 0.1. We use the model's default setting, but the learning rate is set to 5e-5 in the LCL model, 3e-5 (SST2, CR, RT), and 5e-5 (ISEAR, AMAN, TREC) in the Kil model, and 5e-5 in the DualCL model. We implement CoSE using the provided language model\footnote{https://github.com/wangjin0818/CoSE} to generate word embeddings. Under the Kil model comparison, we retain the knowledge-enhanced component in BERT's and Roberta's experiments but exclude them in the CoSE and BERTCK experiments to ensure that the enhancement was derived from a single knowledge source.  
Five repeats of the experiment are conducted and the average classification accuracy results are reported. In each of the repeats, a different seed is used for random data split if no train/dev/test dataset is provided. All the experiments are implemented under Python 3.7 environment and PyTorch 1.10.1. with Cuda version 10.1.

\subsubsection{Classification Results and Analysis}

The results of sentiment analysis are summarized in Table \ref{SAresult}. 
\begin{table*}[htbp]
\caption{The accuracy comparison of models utilizing diverse embeddings for sentiment analysis across various training dataset sizes. 
}
\centering
\large
\resizebox{\linewidth}{!}{
\begin{tabular}{cccccc|ccccc|ccccc}
\hline
                               & \multicolumn{5}{c|}{SST2}                                                                                                                                                                   & \multicolumn{5}{c|}{CR}                                                                                                                                                                                             & \multicolumn{5}{c}{RT}                                                                                                                                                       \\ \cline{2-16} 
\multirow{-2}{*}{Methods}      & 20\%                                                       & 40\%                                   & 60\%                       & 80\%                       & 100\%                       & 20\%                                                       & 40\%                                & 60\%                                & 80\%                                & 100\%                                & 20\%                                   & 40\%                       & 60\%                                & 80\%                                & 100\%                      \\ \hline
\multicolumn{1}{l}{BiLSTM-att} & \multicolumn{1}{l}{}                                       & \multicolumn{1}{l}{}                   & \multicolumn{1}{l}{}       & \multicolumn{1}{l}{}       & \multicolumn{1}{l|}{}       &                                                            & \multicolumn{1}{l}{}                & \multicolumn{1}{l}{}                & \multicolumn{1}{l}{}                & \multicolumn{1}{l|}{}                & \multicolumn{1}{l}{}                   & \multicolumn{1}{l}{}       & \multicolumn{1}{l}{}                & \multicolumn{1}{l}{}                & \multicolumn{1}{l}{}       \\
BERT                           & 0.8364                                                     & \multicolumn{1}{l}{0.8557}             & \multicolumn{1}{l}{0.8654} & \multicolumn{1}{l}{0.8709} & 0.8800                      & \multicolumn{1}{l}{0.7973}                                 & \multicolumn{1}{l}{0.8229}          & \multicolumn{1}{l}{0.8419}          & \multicolumn{1}{l}{0.8552}          & \multicolumn{1}{l|}{0.8577}          & 0.8211                                 & 0.8374                     & 0.8315                              & 0.8452                              & 0.8587                     \\
Roberta                        & \multicolumn{1}{l}{0.8428}                                 & 0.8578                                 & \multicolumn{1}{l}{0.8665} & \multicolumn{1}{l}{0.8740} & \multicolumn{1}{l|}{0.8828} & \multicolumn{1}{l}{0.8010}                                 & \multicolumn{1}{l}{0.8238}          & \multicolumn{1}{l}{0.8453}          & \multicolumn{1}{l}{0.8560}          & \multicolumn{1}{l|}{0.8581}          & 0.8217                                 & 0.8377                     & 0.8467                              & 0.8596                              & 0.8647                     \\
CoSE                           & \multicolumn{1}{l}{{  0.8055}}          & {  0.8531}          & 0.8778                     & 0.8824                     & 0.8874                      & \multicolumn{1}{l}{{ 0.7920}}          & { 0.8187}       & 0.8507                              & 0.8512                              & 0.8640                               & \multicolumn{1}{l}{0.7844}             & \multicolumn{1}{l}{0.8041} & \multicolumn{1}{l}{0.8478}          & \multicolumn{1}{l}{0.8555}          & \multicolumn{1}{l}{0.8613} \\
BERTCK                         & \multicolumn{1}{l}{{  \textbf{0.8621}}} & {  \textbf{0.8780}} & \textbf{0.8835}            & \textbf{0.8868}            & \textbf{0.8923}             & \multicolumn{1}{l}{{  \textbf{0.8261}}} & \textbf{0.8363}                     & \textbf{0.8555}                     & \textbf{0.8667}                     & \textbf{0.8741}                      & {  \textbf{0.8341}} & \textbf{0.8457}            & \textbf{0.8547}                     & \textbf{0.8604}                     & \textbf{0.8697}            \\ \hline
\multicolumn{1}{l}{LCL}        & \multicolumn{1}{l}{}                                       & \multicolumn{1}{l}{}                   & \multicolumn{1}{l}{}       & \multicolumn{1}{l}{}       & \multicolumn{1}{l|}{}       &                                                            &                                     & \textbf{}                           &                                     &                                      & \multicolumn{1}{l}{}                   & \multicolumn{1}{l}{}       & \multicolumn{1}{l}{}                & \multicolumn{1}{l}{}                & \multicolumn{1}{l}{}       \\
BERT                           & 0.8654                                                     & 0.8740                                 & \multicolumn{1}{l}{0.8874} & \multicolumn{1}{l}{0.8984} & 0.9095                      & \multicolumn{1}{l}{0.8643}                                 & \multicolumn{1}{l}{0.8951}          & 0.9073                              & \multicolumn{1}{l}{0.9114}          & 0.9205                               & \multicolumn{1}{l}{0.8481}             & 0.8594                     & 0.8631                              & \multicolumn{1}{l}{0.8753}          & 0.8890                     \\
Roberta                        & 0.8825                                                     & 0.8951                                 & 0.9027                     & 0.9051                     & 0.9111                      & 0.8617                                                     & 0.9043                              & 0.9053                              & 0.9122                              & 0.9308                               & 0.8585                                 & \multicolumn{1}{l}{0.8697} & 0.8725                              & 0.8791                              & 0.8828                     \\
CoSE                           & \multicolumn{1}{l}{0.8227}                                 & \multicolumn{1}{l}{0.8357}             & \multicolumn{1}{l}{0.8933} & \multicolumn{1}{l}{0.9032} & \multicolumn{1}{l|}{0.9148} & \multicolumn{1}{l}{0.8504}                                 & \multicolumn{1}{l}{0.9064}          & \multicolumn{1}{l}{0.9149}          & \multicolumn{1}{l}{0.9202}          & \multicolumn{1}{l|}{0.9311}          & \multicolumn{1}{l}{0.8379}             & \multicolumn{1}{l}{0.8557} & \multicolumn{1}{l}{0.8690}          & \multicolumn{1}{l}{0.8735}          & \multicolumn{1}{l}{0.8837} \\
BERTCK                         & \textbf{0.9003}                                            & \textbf{0.9067}                        & \textbf{0.9133}            & \textbf{0.9198}            & \textbf{0.9213}             & \textbf{0.8734}                                            & \textbf{0.9138}                     & \textbf{0.9160}                     & \textbf{0.9245}                     & \textbf{0.9379}                      & \textbf{0.8632}                        & \textbf{0.8791}            & \textbf{0.8800}                     & \textbf{0.8838}                     & \textbf{0.8847}            \\ \hline
\multicolumn{1}{l}{Kil}        &                                                            &                                        & \textbf{}                  &                            &                             & \textbf{}                                                  &                                     &                                     &                                     &                                      & \multicolumn{1}{l}{}                   & \multicolumn{1}{l}{}       & \multicolumn{1}{l}{}                & \multicolumn{1}{l}{}                & \multicolumn{1}{l}{}       \\
BERT                           & 0.8928                                                     & 0.9027                                 & \multicolumn{1}{l}{0.9106} & \multicolumn{1}{l}{0.9132} & 0.9154                      & 0.8801                                                     & 0.8915                              & 0.9165                              & 0.9223                              & 0.9276                               & 0.8574                                 & 0.8714                     & 0.8761                              & 0.8782                              & 0.8870                     \\
Roberta                        & 0.8971                                                     & 0.9070                                 & 0.9149                     & 0.9196                     & 0.9220                      & 0.8846                                                     & 0.8958                              & 0.9205                              & 0.9271                              & 0.9372                               & \multicolumn{1}{l}{0.8626}             & \multicolumn{1}{l}{0.8723} & 0.8804                              & 0.8832                              & \multicolumn{1}{l}{0.8873} \\
CoSE                           & \multicolumn{1}{l}{0.8814}                                 & \multicolumn{1}{l}{0.8835}             & 0.9006                     & 0.9077                     & 0.9108                      & \multicolumn{1}{l}{0.8833}                                 & \multicolumn{1}{l}{0.8906}          & \multicolumn{1}{l}{0.9197}          & \multicolumn{1}{l}{0.9207}          & \multicolumn{1}{l|}{0.9216}          & \multicolumn{1}{l}{0.8462}             & \multicolumn{1}{l}{0.8550} & \multicolumn{1}{l}{0.8759}          & \multicolumn{1}{l}{0.8838}          & \multicolumn{1}{l}{0.8854} \\
BERTCK                         & \textbf{0.9083}                                            & \textbf{0.9176}                        & \textbf{0.9273}            & \textbf{0.9292}            & \textbf{0.9303}             & \multicolumn{1}{l}{\textbf{0.9096}}                        & \multicolumn{1}{l}{\textbf{0.9176}} & \multicolumn{1}{l}{\textbf{0.9282}} & \multicolumn{1}{l}{\textbf{0.9309}} & \multicolumn{1}{l|}{\textbf{0.9441}} & \textbf{0.8650}                        & \textbf{0.8752}            & \multicolumn{1}{l}{\textbf{0.8815}} & \multicolumn{1}{l}{\textbf{0.8840}} & \textbf{0.8922}            \\ \hline
\multicolumn{1}{l}{DualCl}     &                                                            &                                        &                            &                            &                             &                                                            &                                     &                                     &                                     &                                      & \multicolumn{1}{l}{}                   & \multicolumn{1}{l}{}       & \multicolumn{1}{l}{}                & \multicolumn{1}{l}{}                & \multicolumn{1}{l}{}       \\
BERT                           & 0.8370                                                     & 0.8638                                 & 0.8948                     & 0.9017                     & 0.9106                      & 0.8803                                                     & 0.8989                              & 0.9129                              & 0.9176                              & 0.9237                               & 0.8463                                 & 0.8678                     & 0.8707                              & 0.8763                              & 0.8798                     \\
Roberta                        & 0.8504                                                     & 0.8777                                 & 0.8940                     & 0.9028                     & 0.9149                      & 0.8836                                                     & 0.9096                              & 0.9146                              & 0.9198                              & 0.9286                               & 0.8585                                 & 0.8697                     & 0.8779                              & 0.8845                              & 0.8872                     \\
CoSE                           & 0.8210                                                     & 0.8528                                 & 0.8765                     & 0.8822                     & 0.9080                      & 0.8649                                                     & 0.8775                              & 0.8941                              & 0.9122                              & 0.9229                               & 0.8482                                 & 0.8670                     & 0.8706                              & 0.8828                              & 0.8894                     \\
BERTCK                         & \textbf{0.9066}                                            & \textbf{0.9132}                        & \textbf{0.9149}            & \textbf{0.9154}            & \textbf{0.9289}             & \textbf{0.9043}                                            & \textbf{0.9138}                     & \textbf{0.9229}                     & \textbf{0.9274}                     & \textbf{0.9388}                      & \textbf{0.8665}                        & \textbf{0.8763}            & \textbf{0.8838}                     & \textbf{0.8894}                     & \textbf{0.8971}            \\ \hline
\end{tabular}
}

\label{SAresult}
\end{table*}
Our embedding technique consistently surpasses benchmark embedding methods in performance, demonstrating enhanced accuracy across a variety of datasets and with all tested classification models. To clarify the measurement of effectiveness, we define "average accuracy improvements" as the mean increase in accuracy observed across all compared embedding methods and training set splits within a single classification model. In the SST2 dataset, it achieved average accuracy improvements of 1.79\%, 2.52\%, 1.73\%, and 3.66\% with BiLSTM-ATT, LCL, Kil, and DualCl models, respectively. Similarly, in the CR dataset, the increments in accuracy were 1.60\%, 1.07\%, 1.68\%, and 1.74\%, respectively, and for the RT dataset, the improvements registered were 1.44\%, 1.23\%, 0.61\%, and 1.08\%, respectively. Remarkably, our method proved its efficacy even under constrained training data scenarios. When utilizing only 20\% or 40\% of the available data for training, it demonstrated significant performance, especially compared to CoSE, achieving up to an 8.56\% increase under the DualCl SST2 20\% training setting.

The results of emotion recognition and question answering are summarized in Table \ref{ERQAresult}. 
\begin{table*}[htbp]
\caption{The accuracy comparison of models utilizing diverse embeddings for emotion recognition and question answering across various training dataset sizes. 
}
\centering
\large
\resizebox{\linewidth}{!}{
\begin{tabular}{cccccc|ccccc|ccccc}
\hline
                               & \multicolumn{5}{c|}{ISEAR}                                                                                                                                                    & \multicolumn{5}{c|}{AMAN}                                                                                                                                                           & \multicolumn{5}{c}{TREC}                                                                                                                                                     \\ \cline{2-16} 
\multirow{-2}{*}{Methods}      & 20\%                                   & 40\%                                & 60\%                       & 80\%                                & 100\%                       & 20\%                       & 40\%                                & 60\%                                & 80\%                                & 100\%                                & 20\%                                   & 40\%                                & 60\%                       & 80\%                       & 100\%                               \\ \hline
\multicolumn{1}{l}{BiLSTM-att} & \multicolumn{1}{l}{}                   & \multicolumn{1}{l}{}                & \multicolumn{1}{l}{}       & \multicolumn{1}{l}{}                & \multicolumn{1}{l|}{}       &                            & \multicolumn{1}{l}{}                & \multicolumn{1}{l}{}                & \multicolumn{1}{l}{}                & \multicolumn{1}{l|}{}                & \multicolumn{1}{l}{}                   & \multicolumn{1}{l}{}                & \multicolumn{1}{l}{}       & \multicolumn{1}{l}{}       & \multicolumn{1}{l}{}                \\
BERT                           & 0.4871                                 & 0.5600                              & 0.6037                     & 0.6151                              & 0.6280                      & 0.7034                     & 0.7701                              & 0.7912                              & 0.8054                              & 0.8154                               & 0.9096                                 & 0.9216                              & 0.9424                     & \multicolumn{1}{l}{0.9486} & 0.9536                              \\
Roberta                        & 0.4921                                 & 0.5715                              & 0.6057                     & 0.6190                              & 0.6295                      & 0.7110                     & 0.7716                              & 0.8068                              & 0.8291                              & 0.8318                               & \multicolumn{1}{l}{0.9128}             & \multicolumn{1}{l}{0.9244}          & \multicolumn{1}{l}{0.9412} & 0.9512                     & \multicolumn{1}{l}{0.9560}          \\
BERTCK                        & {  \textbf{0.5098}} & \textbf{0.5812}                     & \textbf{0.6190}            & \textbf{0.6266}                     & \textbf{0.6333}             & \textbf{0.7626}            & \textbf{0.7917}                     & \textbf{0.8154}                     & \textbf{0.8394}                     & \textbf{0.8435}                      & {  \textbf{0.9202}} & \textbf{0.9392}                     & \textbf{0.9540}            & \textbf{0.9616}            & \textbf{0.9668}                     \\ \hline
\multicolumn{1}{l}{LCL}        & \multicolumn{1}{l}{}                   & \multicolumn{1}{l}{}                & \multicolumn{1}{l}{}       & \multicolumn{1}{l}{}                & \multicolumn{1}{l|}{}       &                            &                                     & \textbf{}                           &                                     &                                      & \multicolumn{1}{l}{}                   & \multicolumn{1}{l}{}                & \multicolumn{1}{l}{}       & \multicolumn{1}{l}{}       & \multicolumn{1}{l}{}                \\
BERT                           & 0.6005                                 & 0.6492                              & 0.6596                     & 0.6750                              & 0.6913                      & 0.8081                     & 0.8504                              & 0.8606                              & 0.8638                              & 0.8655                               & 0.9452                                 & 0.9560                              & 0.9624                     & 0.9640                     & \multicolumn{1}{l}{0.9680}          \\
Roberta                        & \multicolumn{1}{l}{0.6109}             & \multicolumn{1}{l}{0.6428}          & \multicolumn{1}{l}{0.6602} & \multicolumn{1}{l}{0.6830}          & \multicolumn{1}{l|}{0.7059} & 0.8147                     & 0.8510                              & 0.8682                              & 0.8699                              & 0.8696                               & \multicolumn{1}{l}{0.9496}             & \multicolumn{1}{l}{0.9592}          & \multicolumn{1}{l}{0.9640} & \multicolumn{1}{l}{0.9648} & 0.9680                              \\
BERTCK                        & \textbf{0.6211}                        & \textbf{0.6574}                     & \textbf{0.6754}            & \textbf{0.6895}                     & \textbf{0.7162}             & \textbf{0.8328}            & \textbf{0.8609}                     & \textbf{0.8709}                     & \textbf{0.8778}                     & \textbf{0.8808}                      & \textbf{0.9600}                        & \textbf{0.9608}                     & \textbf{0.9732}            & \textbf{0.9794}            & \textbf{0.9790}                     \\ \hline
\multicolumn{1}{l}{Kil}        &                                        &                                     & \textbf{}                  &                                     &                             & \textbf{}                  &                                     &                                     &                                     &                                      & \multicolumn{1}{l}{}                   & \multicolumn{1}{l}{}                & \multicolumn{1}{l}{}       & \multicolumn{1}{l}{}       & \multicolumn{1}{l}{}                \\
BERT                           & 0.6257                                 & 0.6627                              & 0.6706                     & 0.6766                              & 0.6857                      & 0.8440                     & 0.8640                              & 0.8733                              & 0.8817                              & 0.8825                               & 0.9520                                 & 0.9676                              & 0.9712                     & 0.9724                     & 0.9744                              \\
Roberta                        & 0.6324                                 & 0.6650                              & \multicolumn{1}{l}{0.6772} & 0.6808                              & 0.6908                      & \multicolumn{1}{l}{0.8545} & 0.8716                              & 0.8814                              & 0.8858                              & 0.8887                               & \multicolumn{1}{l}{0.9608}             & 0.9708                              & \multicolumn{1}{l}{0.9728} & \multicolumn{1}{l}{0.9732} & 0.9748                              \\
BERTCK                        & \multicolumn{1}{l}{\textbf{0.6360}}    & \multicolumn{1}{l}{\textbf{0.6686}} & \textbf{0.6802}            & \multicolumn{1}{l}{\textbf{0.6847}} & \textbf{0.6929}             & \textbf{0.8567}            & \multicolumn{1}{l}{\textbf{0.8753}} & \multicolumn{1}{l}{\textbf{0.8839}} & \multicolumn{1}{l}{\textbf{0.8900}} & \multicolumn{1}{l|}{\textbf{0.8924}} & \textbf{0.9616}                        & \multicolumn{1}{l}{\textbf{0.9716}} & \textbf{0.9736}            & \textbf{0.9740}            & \multicolumn{1}{l}{\textbf{0.9750}} \\ \hline
\multicolumn{1}{l}{DualCl}     &                                        &                                     &                            &                                     &                             &                            &                                     &                                     &                                     &                                      & \multicolumn{1}{l}{}                   & \multicolumn{1}{l}{}                & \multicolumn{1}{l}{}       & \multicolumn{1}{l}{}       & \multicolumn{1}{l}{}                \\
BERT                           & 0.2063                                 & 0.4544                              & \multicolumn{1}{l}{0.6240} & \multicolumn{1}{l}{0.6706}          & 0.6840                      & \multicolumn{1}{l}{0.5012} & \multicolumn{1}{l}{0.6213}          & 0.7196                              & \multicolumn{1}{l}{0.7504}          & 0.7819                               & \multicolumn{1}{l}{0.2596}             & \multicolumn{1}{l}{0.3760}          & \multicolumn{1}{l}{0.7824} & 0.9408                     & 0.9638                              \\
Roberta                        & \multicolumn{1}{l}{0.2375}             & \multicolumn{1}{l}{0.5012}          & 0.6257                     & 0.6770                              & 0.6864                      & 0.5346                     & 0.6548                              & 0.7330                              & 0.7642                              & 0.7858                               & 0.3120                                 & 0.4120                              & 0.9384                     & 0.9432                     & 0.9736                              \\
BERTCK                        & \textbf{0.6408}                        & \textbf{0.6705}                     & \textbf{0.6761}            & \textbf{0.6844}                     & \textbf{0.6940}             & \textbf{0.7161}            & \textbf{0.7262}                     & \textbf{0.7543}                     & \textbf{0.7885}                     & \textbf{0.8020}                      & \textbf{0.9488}                        & \textbf{0.9496}                     & \textbf{0.9584}            & \textbf{0.9664}            & \textbf{0.9755}                     \\ \hline
\end{tabular}
}

\label{ERQAresult}
\end{table*}
In fine-grained classification tasks, our method demonstrates a significant accuracy improvement over other embeddings, especially with smaller training sets. In the BiLSTM-ATT model, we observed an average accuracy enhancement of 1.28\%, 2.69\%, and 1.22\% in ISEAR, AMAN, and TREC datasets, respectively. Similarly, the LCL model yielded improvements of 1.41\%, 1.25\%, and 1.04\% in the same datasets. Although the Kil model, a learning-based knowledge-enhanced model, doesn't showcase as pronounced improvements, the scenario was markedly different for the complex deep-learning DualCl model. DualCl, which typically struggles with learning from limited data, exhibited remarkable performance boosts with our knowledge-based embedding. Under the 20\% setting, accuracy surged by 41.89\%, 19.82\%, and 66.3\% for ISEAR, AMAN, and TREC respectively. The trend persisted in the 40\% setting, with gains of 19.27\%, 8.82\%, and 55.56\% recorded for the same datasets.

Upon the evaluation of three classification applications, it is concluded that the models incorporating knowledge-based embeddings outperform their original ones in accuracy, particularly when constrained by smaller training datasets.

\subsubsection{Extension of Knowledge-based Embedding Learning to GloVe}

 \begin{table*}[htbp]
\caption{The accuracy comparison of models utilizing diverse embeddings for text classifications based on GloVe. 
}
\centering
\resizebox{\linewidth}{!}{
\begin{tabular}{cccccc|ccccc|ccccc}
\hline
                               & \multicolumn{5}{c|}{CR}                                                                                                                                                                     & \multicolumn{5}{c|}{ISEAR}                                                                                                          & \multicolumn{5}{c}{TREC}                                                                                                           \\ \cline{2-16} 
\multirow{-2}{*}{Methods}      & 20\%                                                       & 40\%                                   & 60\%                       & 80\%                       & 100\%                       & 20\%                                   & 40\%                 & 60\%                 & 80\%                 & 100\%                 & 20\%                                   & 40\%                 & 60\%                 & 80\%                 & 100\%                \\ \hline
\multicolumn{1}{l}{BiLSTM-att} & \multicolumn{1}{l}{}                                       & \multicolumn{1}{l}{}                   & \multicolumn{1}{l}{}       & \multicolumn{1}{l}{}       & \multicolumn{1}{l|}{}       &                                        & \multicolumn{1}{l}{} & \multicolumn{1}{l}{} & \multicolumn{1}{l}{} & \multicolumn{1}{l|}{} & \multicolumn{1}{l}{}                   & \multicolumn{1}{l}{} & \multicolumn{1}{l}{} & \multicolumn{1}{l}{} & \multicolumn{1}{l}{} \\
GloVe                          & \multicolumn{1}{l}{0.6928}                                 & \multicolumn{1}{l}{0.7125}             & \multicolumn{1}{l}{0.7483} & \multicolumn{1}{l}{0.7653} & \multicolumn{1}{l|}{0.7853} & 0.3243                                 & 0.3863               & 0.6170               & 0.6656               & 0.6873                & 0.7740                                 & 0.8211               & 0.8384               & 0.8562               & 0.8676               \\
GloVeCK                        & \multicolumn{1}{l}{{  \textbf{0.7147}}} & {  \textbf{0.7408}} & \textbf{0.7579}            & \textbf{0.7733}            & \textbf{0.7920}             & {  \textbf{0.4240}} & \textbf{0.4710}      & \textbf{0.6399}      & \textbf{0.6967}      & \textbf{0.7071}       & {  \textbf{0.8420}} & \textbf{0.8660}      & \textbf{0.8800}      & \textbf{0.9000}      & \textbf{0.9040}      \\ \hline
\multicolumn{1}{l}{Kil}        &                                                            &                                        & \textbf{}                  &                            &                             & \textbf{}                              &                      &                      &                      &                       & \multicolumn{1}{l}{}                   & \multicolumn{1}{l}{} & \multicolumn{1}{l}{} & \multicolumn{1}{l}{} & \multicolumn{1}{l}{} \\
GloVe                          & 0.8263                                                     & 0.8264                                 & 0.8417                     & 0.8420                     & 0.8538                      & 0.2578                                 & 0.2946               & 0.5386               & 0.6195               & 0.6266                & 0.7255                                 & 0.7857               & 0.7956               & 0.8224               & 0.8337               \\
GloVeCK                        & \textbf{0.8346}                                            & \textbf{0.8341}                        & \textbf{0.8437}            & \textbf{0.8550}            & \textbf{0.8650}             & \textbf{0.2833}                        & \textbf{0.3048}      & \textbf{0.5627}      & \textbf{0.6248}      & \textbf{0.6330}       & \textbf{0.7996}                        & \textbf{0.8257}      & \textbf{0.8377}      & \textbf{0.8497}      & \textbf{0.8597}      \\ \hline
\end{tabular}
}

\label{gloveresult}

\end{table*}

To test whether the proposed lexical knowledge-based embedding learning can be extended to other embedding learning models besides BERT, we conducted experiments on GloVe embedding \citep{pennington2014glove} for sentiment analysis. Our knowledge-based embedding learning model maps GloVe word embedding into the discriminative space (denoted as GloVeCK). The 300-dimensional GloVe embedding is adopted here. The lexical knowledge used by the GloVe-based embedding learning model is similar to that of BERT but with certain sub-pieces removed from GloVe's vocabulary. Additionally, the deployment of the learning embedding is the same as that of BERT's. We evaluated the performance of the model on three text classification tasks: CR, ISEAR, and TREC based on BiLSTM-att and Kil. The learning rates used in the experiments were set to 9e-4, except for TREC in BiLSTM-att, where the learning rate was set to 5e-4.

The results in Table \ref{gloveresult} prove that the proposed knowledge-based embedding learning method is equally applicable to GloVe word embedding. Our approach has resulted in improved performance across all datasets. Notably, the improvement was particularly significant under small settings such as 20\% and 40\% settings, with the best performance improvement ranging from 2.83\% (BiLSTM-att 40\% setting in CR), 7.41\% (Kil 20\% setting in TREC) to 9.97\% (BiLSTM-att 20\% setting in ISEAR).

\section{Conclusion}
This paper presents a lexical knowledge-based word embedding learning method. This method projects pre-trained word embedding into more discriminative embeddings with maximized within-class similarity and between-class difference. The new knowledge-based embedding learning method has two advantages. First, more discriminative word embedding facilitates the subsequent classification task. Second, the new method works on pre-trained embeddings without requiring re-training or fine-tuning of the embedding learning model, such as BERT or Glove, and is therefore computationally efficient. The proposed method is applicable to any text classification task as long as a domain-specific lexicon exists. If the lexicon is not readily available, it can be acquired from online open sources using the proposed lexical knowledge acquisition algorithm. One limitation of utilizing a knowledge base is the reliability of domain knowledge collected from open resources. Although advanced searching tools are employed, lexicon overlapping between different classes still occurs. To filter out overlapping words, additional post-processing is needed. In our future work, we will investigate the potential of leveraging ChatGPT or other large language models to assist lexicon construction.

\section*{Acknowledgements}
The research was conducted at the Future Resilient Systems at the Singapore-ETH Centre and was supported by the National Research Foundation Singapore under its Campus for Research Excellence and Technological Enterprise program.

\bibliography{acl_latex}

\end{document}